\documentclass[10pt,twocolumn,letterpaper]{article}

\usepackage{cvpr}              
\usepackage{graphicx}
\usepackage{svg}
\usepackage[linesnumbered,ruled,vlined]{algorithm2e}
\usepackage{multirow} 
\usepackage[misc]{ifsym}
\definecolor{cvprblue}{rgb}{0.21,0.49,0.74}
\usepackage[pagebackref,breaklinks,colorlinks,citecolor=cvprblue]{hyperref}

\def\paperID{9591} 
\def\confName{CVPR}
\def\confYear{2024}

\title{M$^{2}$Chat: Empowering VLM for Multimodal LLM Interleaved \\Text-Image Generation}

\author{Xiaowei Chi$^{1,\dag}$, Junbo Qi$^{2,\dag}$, Rongyu Zhang$^{3,\dag }$, Shanghang Zhang$^{3}$, Qifeng Liu$^{1, \textrm{\Letter}}$, Yike Guo$^{1, \textrm{\Letter}}$\\
$^{1}$ The Hong Kong University of Science and Technology\\   
$^{2}$ Waseda University,
$^{3}$ Peking University\\
 {\tt\small xchiaa@connect.ust.hk, \{liuqifeng, yikeguo\}@ust.hk } \\
}

\begin{document}
\twocolumn[{%
\renewcommand\twocolumn[1][]{#1}%
\maketitle
\vspace{-10mm}
\begin{center}
    \centering
    \includegraphics[width=\linewidth]{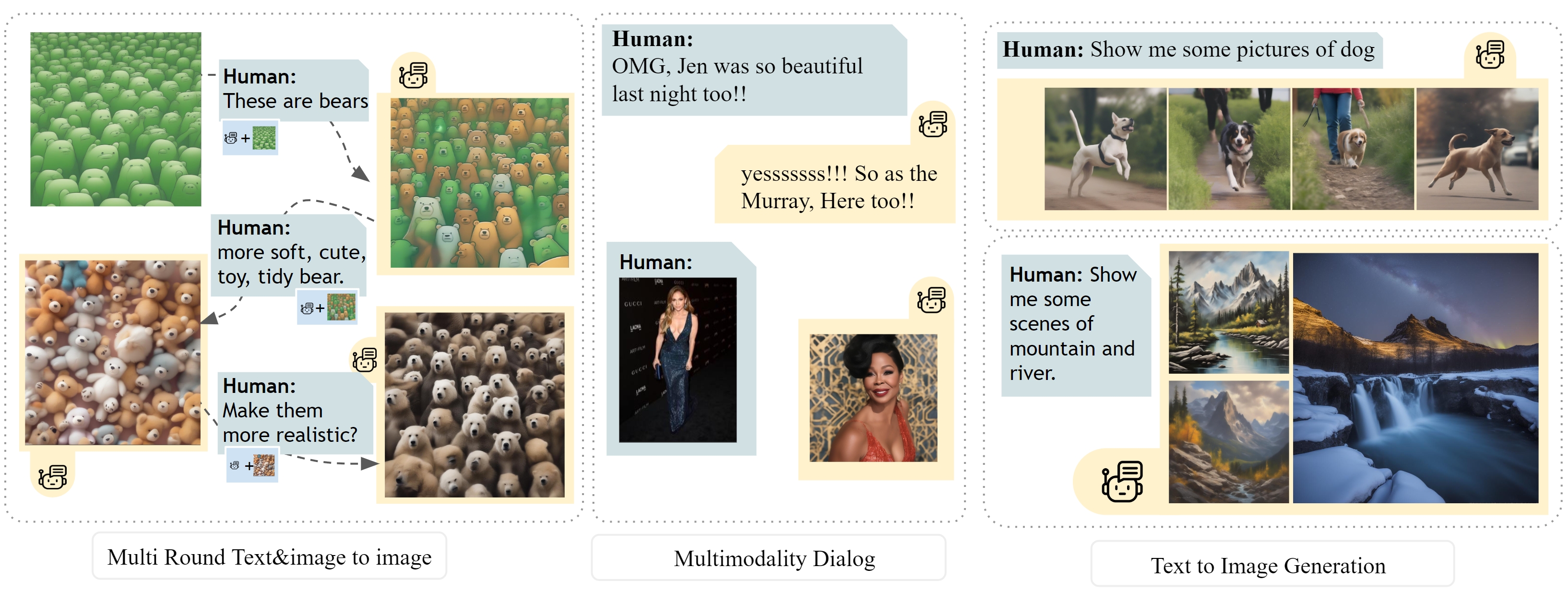}
    \vspace{-2em}
    \captionof{figure}{Advanced capabilities of our proposed M$^{2}$Chat in interleaved multimodal chat, multi-round text and image-to-image generation, and text-to-image generation. }
    \label{fig:teaser}
\end{center}
}]

\maketitle
\begin{abstract}
While current LLM chatbots like GPT-4V bridge the gap between human instructions and visual representations to enable text-image generations, they still lack efficient alignment methods for high-fidelity performance on multiple downstream tasks. In this paper, we propose \textbf{$M^{2}Chat$}, a novel unified multimodal LLM framework for generating interleaved text-image conversation across various scenarios. Specifically, we propose an $M^{3}Adapter$ that efficiently integrates granular low-level visual information and high-level semantic features from multi-modality prompts. Upon the well-aligned fused feature, $M^{3}Adapter$ tailors a learnable gating strategy to balance the model creativity and consistency across various tasks adaptively. Moreover, to further enhance the effectiveness of $M^{3}Adapter$ while preserving the coherence of semantic context comprehension, we introduce a two-stage $M^{3}FT$ fine-tuning strategy. This strategy optimizes disjoint groups of parameters for image-text alignment and visual-instruction respectively. Extensive experiments demonstrate our $M^{2}Chat$ surpasses state-of-the-art counterparts across diverse benchmarks, showcasing its prowess in interleaving generation, storytelling, and multimodal dialogue systems. The demo and code are available at 
\href{https://mattie-e.github.io/M2Chat.github.io}{https://mattie-e.github.io/M2Chat.github.io}.

\end{abstract}    
\renewcommand{\thefootnote}{\fnsymbol{footnote}}
\footnotetext{$\textrm{\Letter}$ Corresponding authors,   $\textrm{\dag}$ Equal contribution.}
  \section{Introduction}
\label{sec:intro}
In the realm of burgeoning large-scale vision-and-language models (VLMs), the integration of multi-modal features represents more than a mere trend; it is a pivotal breakthrough that is sculpting an extensive range of applications, including object detection~\cite{wang2023cogvlm, lin2023sphinx}, Optical Character Recognition (OCR)~\cite{liu2023hidden}, and Visual-Question-Answering (VQA)~\cite{liu2023llava, liu2023improvedllava, zhang2023LLaMAadapter,zhu2023minigpt4,gao2023llamaadapterv2, lin2023sphinx, wang2023cogvlm}. In light of the escalating demand for human-machine chat applications across numerous domains, such as virtual reality, social media, and e-commerce, there is heightened anticipation for VLMs to adeptly interpret and synthesize multi-modality content cohesively for substantially enhancing the quality of conversations. Nevertheless, prevailing research such as MiniGPT-5~\cite{zheng2023minigpt5} and DreamLLM~\cite{dong2023dreamllm} has concentrated predominantly on refining the multi-modal alignment ~\cite{qi2023weakly} and interleaving generalization capabilities to enhance performance in tasks like image-editing and long-context generation. However, previous approaches uniformly apply the same knowledge across various tasks, neglecting to account for the task-specific inherent characteristics of VLMs.

As evidenced in previous works, considering employing the VLM on various downstream tasks while preserving coherent semantic comprehension, there are still two challenges: 
1) Since the vast and intricately complex multi-modality features from various downstream tasks, it is quite difficult to obtain aligned coherent text-image pairs in a unified space effectively. 2) Directly applying the visual language model is not adequately tailored for modeling the diverse and contextually consistent text-image dialogue from the unified space.

To address the challenges outlined, we introduce $M^{2}Chat$, an innovative model for interleaved multimodal generation. $M^{2}Chat$ adepts at creating text-image pairs that are both contextually consistent and creatively imaginative, tailored with relevant knowledge for diverse tasks. Specifically, by integrating Stable Diffusion XL\cite{podell2023sdxl} with LLaMA-AdapterV2\cite{gao2023llamaadapterv2}, we developed a task-specific Multimodal Multi-level Adapter ($M^{3}Adapter$). This adapter efficiently integrates low-level visual information and high-level semantic features from multimodality prompts through a learnable gating strategy, effectively balancing the contributions of each modality. This approach maintains a delicate equilibrium in the M$^{3}$Chat to balance consistency with incongruity towards diverse tasks. 

Meanwhile, we further devised a two-stage Multimodal Mixed Fine-Tuning strategy, denoted as $M^{3}FT$, which strategically optimizes distinct sets of parameters tailored specifically for image-text alignment and visual-instruction tasks. In the first stage, we finetune the parameter groups for alignment to projected the multimodal features with the input dimension of the image generation model. Then, in the second stage, we tailored a specific token and further training the $M^{3}Adapter$ components with instruction data from different fields.

Empirical evidence highlights M$^{2}$Chat's superior capabilities in tasks like image editing, storytelling, and multimodal dialogue, outperforming current models in fine-tuning efficiency and generation quality, with a proficiency in creating imaginary but coherent images and text. The contributions of our study are outlined as follows:
\begin{itemize}
    \item We have developed \textit{$M^{2}Chat$}, which is an innovative VLM capable of seamless text-image interleaved generation across a range of tasks, especially on complex multimodal dialogue scenarios.
    \item The $M^{3}Adapter$ aligns VLM with Stable Diffusion XL for enhanced multimodal fusion, using an adaptive gate for multi-level feature integration, ensuring generation creative-consistency balance for diverse tasks.
    \item We further design a two-stage tuning strategy M$^{3}$FT that cooperates with M$^{3}$Adapter to align text and image while maintaining semantic coherence.
\end{itemize}
\section{Related Work}
\label{sec:relatework}

\subsection{Multimodal Large Language Model}
Researchers in the field of multimodal large language models have devoted considerable attention to image understanding. Several studies have specifically focused on learning captioning abilities using text-image paired datasets, such as KOSMOS-1~\cite{huang2023language}, FROMAGe~\cite{koh2023grounding}, and BLIP-2~\cite{li2023blip}. While the majority of research has concentrated on computer vision and LLM, there has also been attention given to improving the fine-tuning capabilities of instructing models like Llava~\cite{liu2023llava}, Llava1.5~\cite{liu2023improvedllava}, and MiniGPT4~\cite{zhu2023minigpt4}. 
Moreover, recently, the visual understanding model has achieved impressive improvement. Opensource model like LlaVA-NeXT \cite{liu2024llava} integrate the multiple visual understanding tasks, including object detection and OCR, so as SPHINX\cite{lin2023sphinx}. Some efforts have aimed to incorporate more modalities, as demonstrated in works like Video-LLaMA~\cite{damonlpsg2023videoLLaMA}. Or, aims at long context movie understanding, like MovieChat\cite{song2023moviechat}. However, few of them noticed the similarity between the VLM and the text-to-image generation task, especially the datasets. The existing works are professional in adding the modality of input to expand the understanding ability of the VLM but ignore expanding the modality of output.

\subsection{VLM Downstream Tasks}
\textbf{Image Generation and Editing.}
The state-of-the-art generation model has shifted from GAN-based approaches to stable diffusion, as highlighted in the work by~\cite{nichol2021improved} and Song~\cite{song2020score}. While stable diffusion is renowned for its strong and controllable image generation capabilities, as proposed by SDXL~\cite{podell2023sdxl}, other works have explored the editing problem in image generation by manipulating the input prompts, as seen in the studies by Cao~\cite{cao2023masactrl} and Hertz~\cite{hertz2022prompt}. Additionally, Zhang~\cite{zhang2023adding} introduced the concept of adding Controlnet to the diffusion model, which enhances the controllability of diffusion-based image generation.

\begin{figure*}[t]
  \centering
  \includegraphics[width=\linewidth]{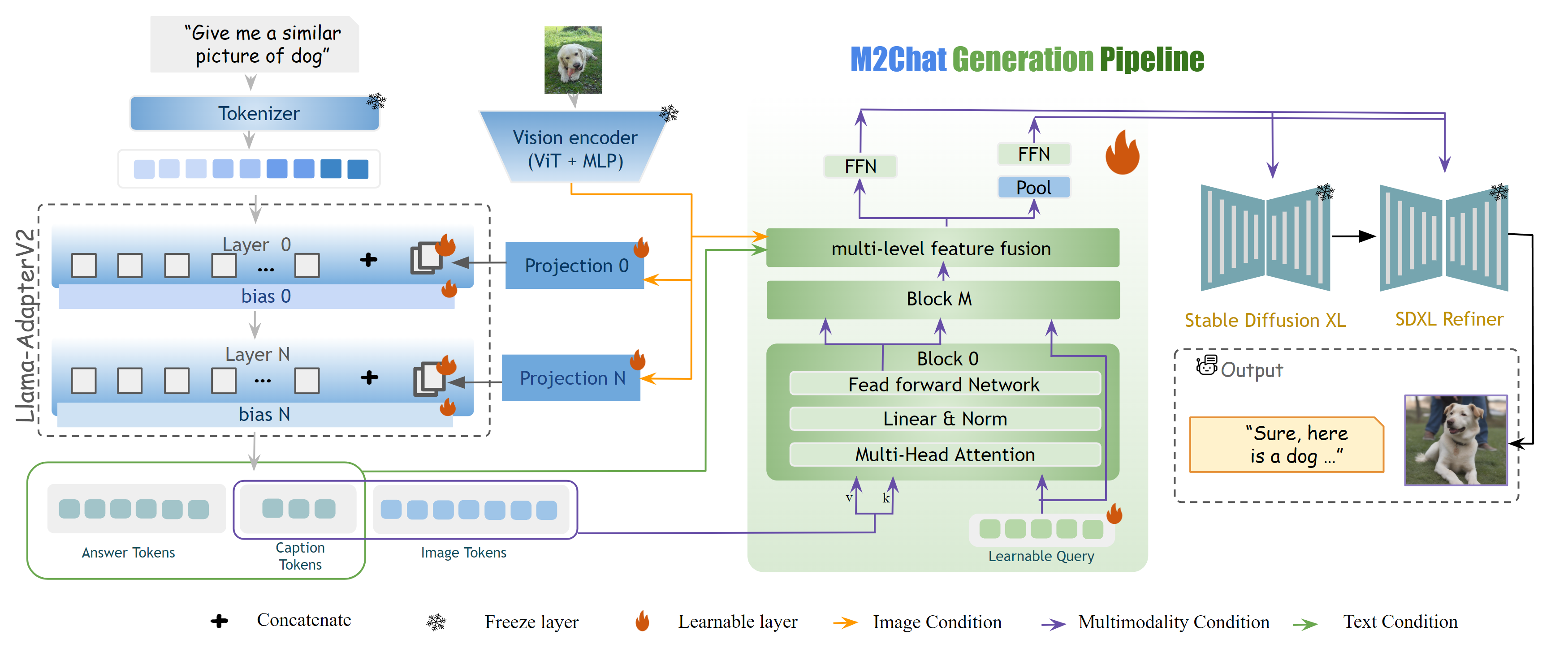}
  \caption{Illustration of $M^{2}Chat$, which features a generation pipeline that processes both image and text inputs, harnessing the capabilities of LLaMA-AdapterV2 \cite{gao2023llamaadapterv2} and SDXL \cite{podell2023sdxl} to craft high-fidelity image-text pairs. Our system excels in three key areas: Text-to-Image (T2I) generation, Storytelling, and Multimodal dialogue. Image generation occurs as VLM forward propagation yields hidden embeddings, which are then utilized to train the M$^{3}$Adapter—distinguished by its minimal trainable parameters.
  }
  \label{fig:framework}
  \vspace{-0.2cm}
\end{figure*}

\noindent\textbf{Interleaving Generation.}
Recent research has explored various approaches to integrate Multimodal Language Models (VLM) with text-image generation tasks. DALLE-3~\cite{openai2023dalle3} relies on prompts for generation without image conditions, while Emu~\cite{sun2023generative} fine-tunes VLM for multimodal context generation but requires substantial computational resources. NextGPT~\cite{wu2023nextgpt} aligns audio, text, and image modalities using adapters but only aligns output text embeddings. DreamLLM~\cite{dong2023dreamllm} generates images and text simultaneously but also requires significant computational resources. MiniDALLE3~\cite{minidalle3} proposes a two-stage generation method using a chat editing model, and SEED-LLaMA~\cite{ge2023seed1,ge2023seed2} aligns LLaMA and generation models with discrete vision tokens. Additionally, chat editing models for 3D models, such as 3D-GPT~\cite{sun20233dgpt}, show promise in this area. Moreover, there are also a lot of explorations of multi-modality generation~\cite{tang2023codi,koh2023generating,qu2023layoutllm,lian2023llm,zhang2023unimodal}. Despite these efforts, efficient alignment and the full exploration of VLM's generalization ability in text-image interleaved generation remain unexplored.

\section{Proposed Method}
\label{method}

In this work, we introduce M$^{2}$Chat, a model that aligns LLaMA-AdapterV2\cite{gao2023llamaadapterv2} with Stable Diffusion XL\cite{podell2023sdxl} for simultaneous text-image generation across diverse tasks. This part is structured as follows. We first introduce the overarching architecture of our framework, including how we construct the visual instruction, the innovative M$^{3}$Adapter, and its custom-designed adaptive gate. We then illustrate the advanced two-stage M$^{3}$FT fine-tuning approach that significantly elevates the generative quality with the multimodal dual-loss objective function

\subsection{Preliminary}
Confronted with the complexities of generating multimodal dialogues with asynchronously aligned image and text semantics, our novel pipeline, depicted in Fig. \ref{fig:framework}, leverages the vision-language model LLaMA-AdapterV2 $\theta_{vlm}$\cite{gao2023llamaadapterv2} to synergize with SDXL $\theta_{sdxl}$\cite{podell2023sdxl}. This orchestrates the generation of cohesive text-image conversations. Particularly, we utilize the VLM as a multimodal encoder and integrate a bespoke M$^{3}$Adapter for aligning multimodal features, thereby streamlining the fusion of text and image narratives, while SDXL facilitates the actual image synthesis.

\paragraph{Visual Instruction Formatting.}
\label{sec:vic}
We begin by detailing our instruction design process. We draw from an image-text dataset $\mathcal{D}: \{\mathcal{X, Y}\}$, containing pairs of images $\{x\}_{i=1}^{N}$ and their corresponding textual contexts $\{y\}_{i=1}^{N}$, where $N$ is the sample count. To construct the context $Y$, we adopt the principles of visual instruction tuning \cite{liu2024visual} and introduce an additional image token $<|img|>$ to denote padding, alongside $<|IC|>$ to signal the start of an image caption. These tokens serve as markers to differentiate token types during the two-stage M$^{3}$FT training phase. 
The context $y$ is organized as 
\begin{equation}
\begin{aligned}y = \{Instruction:& T  \\
Input:& \hat y, \\
Response:& \ \{y_{res}<|IC|>\hat y_{cap}<|img|>\}\}
\end{aligned}
\end{equation}
where $T$ signifies task description prompts, $y_{cap}$ is the initial image caption, $\hat y$ is the preceding dialogue text, and $y_{res}$ is the response context. Moreover, $\hat y_{cap}$ represents the second-round image caption. Our objective is to craft a $\hat y_{cap}$ aligning with the task demands and to produce an image that satisfies the $\hat y_{cap}$.

\subsection{Framework Architecture}
\label{sec:of}


\paragraph{VLM Encoder.}
In anticipation of the multimodal language model's strong text-image encoding capabilities, we incorporate LLaMA-AdapterV2 as our foundational pre-trained VLM. As illustrated in Fig. \ref{fig:framework}, each context in the sequence $\{y\}_{i=1}^{N}$ is encoded into text embeddings $e_{text} \in \mathbb{R}^{length \times 4096}$ using a text encoder. Simultaneously, the corresponding set of images $\{x\}_{i=1}^{N}$ undergo encoding by a visual encoder, yielding visual features $f_{img} \in \mathbb{R}^{length \times 768}$, leveraging a CLIP based ViT+MLP framework \cite{radford2021clip}.

\paragraph{Text-Image Token Generation.}
The VLM outputs a sequence of hidden tokens $t_{out} \in \mathbb{R}^{length \times 4096}$, mirroring the dimension of the input text embeddings. We partition this output into three segments: answer tokens $t_{ans} \in \mathbb{R}^{length_{ans} \times 4096}$, caption tokens $t_{cap}$, and image tokens $t_{img} \in \mathbb{R}^{length_{img} \times 4096}$, with the intricate token structure discussed in the instruction formatting section. The answer tokens $t_{cap}  \in \mathbb{R}^{length_{cap} \times 4096}$ are transformed into human-readable text by LLaMA text decoder, while the image generation tokens $t_{\{cap, img\}}  \in \mathbb{R}^{(length_{cap}+length_{img}) \times 4096}$ provide the foundational features for synthesizing images.

\paragraph{Multimodal Multi-level Adapter: }
\label{sec:Adapter}
To overcome the challenges posed by SDXL's restricted token capacity in representing image-text interactions, which tends to result in inconsistencies across diverse generation tasks, we introduce the Multimodal Multi-level Adapter (M$^{3}$Adapter), denoted as $\theta_{m^{3}a}$. This adapter is seamlessly integrated with the image decoder to deliver uniform and superior-quality outputs. Comprising predominantly of cross-attention and linear layers, the M$^{3}$Adapter's functionality is harnessed through the cross-attention layer:
$Attention(Q,K,V)=softmax(\frac{QK^{T}}{\sqrt{dim}})\cdot V$ with $Q=\mathcal{W}^{(i)}_{Q}\cdot query,
K=\mathcal{W}^{(i)}_{K}\cdot h_{l},
V=\mathcal{W}^{(i)}_{V}\cdot h_{l}$
where $\mathcal{W}^{(i)}_{Q}$, $\mathcal{W}^{(i)}_{K}$, and $\mathcal{W}^{(i)}_{V}$ are learnable attention matrices, with $i$ denoting specific layer in use. 
Specifically, M$^{3}$Adapter transforms VLM outputs $h_{0}=t_{\{cap,img\}}$ into alignment features $h_{align,l}\in\mathbb{R}^{77\times2048}$ and $h_{palign,l}\in\mathbb{R}^{1\times1280}$, which are matched with SDXL text encoder outputs $e_{clip}\in\mathbb{R}^{77\times2048}$ and $e_{pclip}\in\mathbb{R}^{1\times1280}$ using Mean Squared Error (MSE) loss, as demonstrated in the following equation:
\begin{equation}
\begin{aligned}
\label{eq:m^{3}a}
\mathcal{L}_{align} = (h_{palign,l}-e_{pclip})^{2} + \frac{1}{77}\sum^{77}_{k=1}(h^{(k)}_{align,l}-e^{(k)}_{clip})
\end{aligned}
\end{equation}
where $l$ indicates the $l^{th}$ layer. However, aligning features from two modalities directly limits model creativity, particularly in multimodal conversation. Therefore, we introduce a multi-level feature fusion strategy that incorporates granular low-level visual primitive $f_{img}$ encoding layouts and textures into semantically rich deep high-level multimodal feature $h_{l}$ to obtain the final fused feature $f_{fus}$. A learnable gate modulates this fusion for adaptability. This method enriches image generation guidance, tailoring output to varied task demands. We differentiate answer tokens $t_{ans}$ and caption tokens $t_{cap}$ from VLM output $h_{l}$ with special tokens $<|IC|>$, computing their cosine similarity, which acts as a task-specific flag signaling when answers and captions diverge or require consistency in different tasks. We utilize this metric to dynamically adjust the gate's fusion rate of multi-level features for varied scenarios as
\begin{equation}
\begin{aligned}
\label{eq:loss}
f_{fus} = (1-\frac{e_{ans} \cdot e_{cap}}{\| e_{ans} \| \| e_{cap} \|}) \times f_{img} + \frac{e_{ans} \cdot e_{cap}}{\| e_{ans} \| \| e_{cap} \|}\times h_{l}
\end{aligned}
\end{equation}

Such adaptive multi-level feature fusion facilitates resilient image generation with balancing in the creativity of M$^{2}$Chat for multimodal dialogue generation and the model coherence for other interleaved generation tasks, seamlessly accommodating significant text-image variances.

\subsection{Training Strategy}
\label{sec:ft}
\paragraph{First Stage in M$^{3}$FT for Alignment.}
We initially fine-tune the model to ensure multimodal feature alignment contributed to M${3}$Adapter. 
For the synthesis of pertinent visuals, the extracted mapping feature $h_{align}$ and $h_{palign}$ serves as a conditional input during the denoising phase. Drawing parallels with Score Distillation Sampling (SDS)~\cite{Song2020ScoreBasedGM}, we feed aligned text embeddings into SDXL's frozen pre-trained UNet $\theta_{unet}$ to generate a score reflecting the target image distribution which can me mathematically formulated as:
\begin{equation}
\begin{aligned}
\label{eq:m^{3}b}
h_{unet}=\theta_{unet}(\delta_{noise}(\mathcal{I},\lambda),h_{align},h_{palign},\lambda)
\end{aligned}
\end{equation}
where $\mathcal{I}$ is the image feature from the VAE encoder, which is modified by adding $\lambda$ times the random noise $\epsilon$ through $\delta_{noise}(\cdot)$. The calculation of DDPM loss formalized mathematically as
\begin{equation}
\begin{aligned}
\label{eq:loss1}
\mathcal{L}_{ddpm}:= \mathbb{E}_{\epsilon\sim\mathcal{N}(0,1),\lambda}\left [||\epsilon - h_{unet}||^{2}\right ]
\end{aligned}
\end{equation}
With a focus on multimodal alignment, as indicated by the purple line in Fig.~\ref{fig:framework}, we keep the LLaMA-AdapterV2's projection and bias frozen. 
Due to inconsistencies in feature dimensions, model decoder, and even vocabulary, there is a huge domain difference between VLM $\theta_{vlm}$ and SDXL $\theta_{sdxl}$. Therefore, we first apply the alignment loss $\mathcal{L}_{align}$ to complement the original DDPM loss $\mathcal{L}_{ddpm}$ to enhance the general generation quality and formulate the multimodal loss as
\begin{equation}
\begin{aligned}
\label{eq:m^{3}c}
\mathcal{L}_{M^{2}FT} = \mathcal{L}_{ddpm} + \varphi\cdot\mathcal{L}_{align}
\end{aligned}
\end{equation}
where $\varphi$ is a hyperparameter. Note that only the M$^{3}$Adapter undergoes updates during the initial M$^{3}$FT stage. Our model aligns the VLM feature space with SDXL, achieving success in diverse multimodal generation tasks. We provide in-depth visualization and CLIP performance post-first stage training in Sec.\ref{sec:exp}. 

\paragraph{Second Stage in M$^{3}$FT for Consistency.}
For Multimodal Mixed Fine-Tuning (M$^{3}$FT), the target is to tune the model and generate the answer and the image tokens. Since the complexity of MMDialog, the answer and the image have inconsistency in their meaning. In M$^{3}$FT the LLM is tuned by the loss group and DDPM at the same time. We separate the answer token and the caption tokens, tuning the model on the text-image to text-image patterns. In this round, we tune all components of the M3Adapter, including the bias of the LLaMA, the projection of visual tokens, the M$^{3}$FT factor, and the adapters. As shown in the pipeline, each component would be affected multiple times of differences, which would speed up the training process, and efficiently align the components. The overall optimistic function of M$^{3}$FT is as follows
\begin{equation}
\begin{aligned}
\label{eq:m^{3}d}
\mathcal{L}_{M^3FT} = \mathcal{L}_{ddpm} + \varphi\cdot\mathcal{L}_{align} + \cdot\mathcal{L}_{text}
\end{aligned}
\end{equation}
where $\mathcal{L}_{text}$ represents the text conditioning loss, assessing the discrepancy between generated tokens and labels.

\section{Experiments}
\label{sec:exp}

In this section, we analyze and evaluate the generation performance of \textit{$M^{2}Chat$} and the efficiency of M$^{3}$Adapter and M$^{3}$FT across various tasks. The empirical results demonstrate the superiority of our proposed methods against other state-of-the-art baselines in generation quality and semantic consistency.


\subsection{Downstream Tasks}
\label{sec:tasks}


Our paper enhances multimodal LLMs for interleaved generation tasks, producing related and intertwined text and images. Specifically, the interleaving generation task can be defined into several sub-tasks:
\begin{itemize}
    \item \textbf{Chat-based image generation} requires the model to discern and react on often vague user inputs, extracting key elements to produce diverse images that match user intent, showcasing both comprehension and creative alignment with user specifications.
    
    \item \textbf{Interleaving generation} aims to perform basic editing operations based on text instructions. During the editing process, the model emerges with the ability to comprehend human commands and make appropriate editing based on the understanding.
    
    \item \textbf{storytelling} requires the model to weave a coherent narrative with corresponding images, ensuring each image reflects the unfolding story. This demands a deep understanding of context and the ability to create rich text and visuals, delivering an immersive narrative experience.

    \item \textbf{Multimodal Dialogue} diverges from traditional ones by tackling inconsistencies in text-image pairs. VLM must go beyond describing images to generating relevant dialogues and topic-specific visuals, enriching the conversation with images more than content visualization.
\end{itemize}




\begin{figure*}[t]
  \centering
    \includegraphics[width=0.99\linewidth]{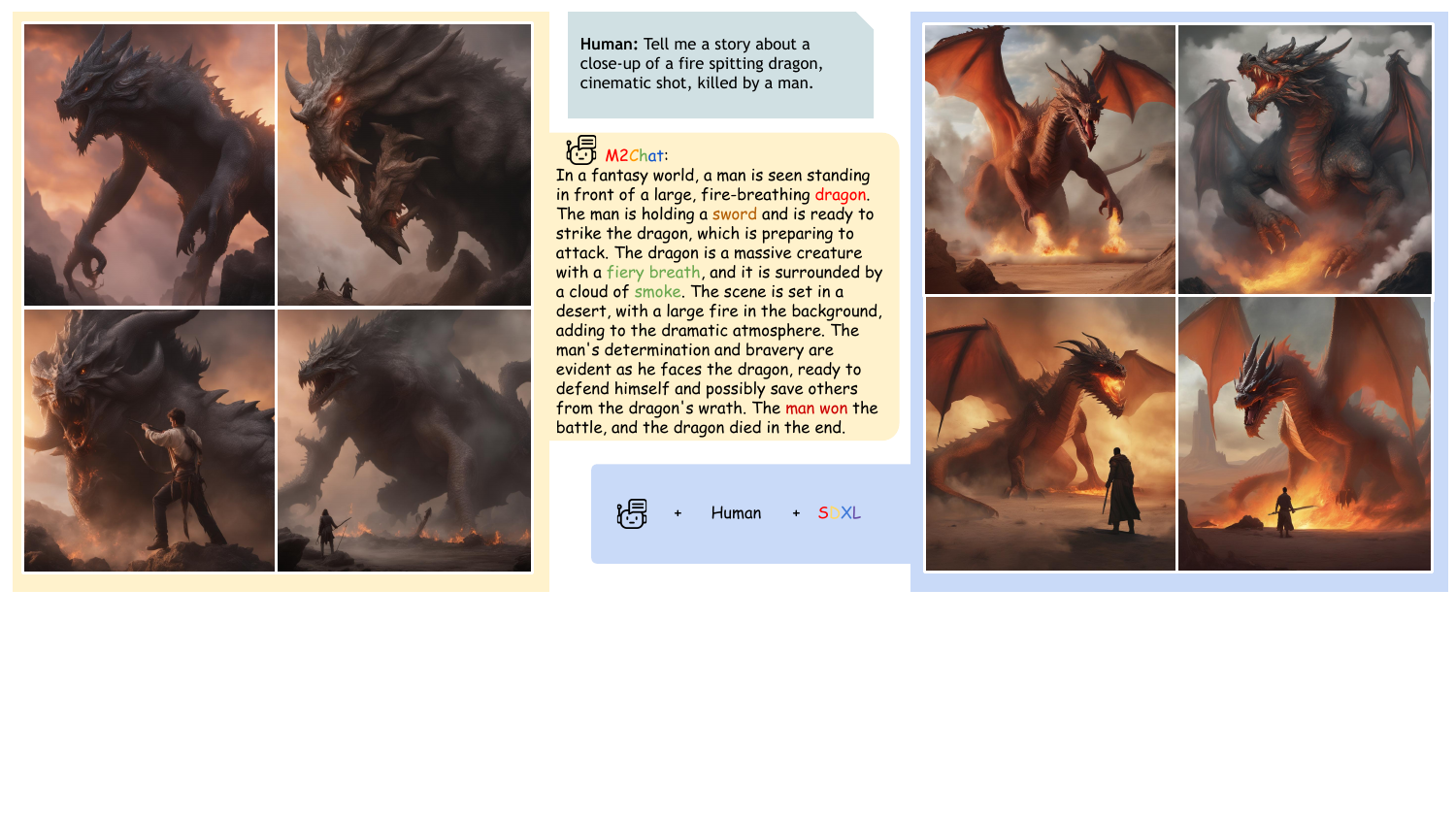}
  \caption{The storytelling pipeline involves the generation of four pictures and a corresponding text story. In this particular example, the human initiates a request to generate a story, starting with the first sentence about a dragon. $M^{2}Chat$ can generate pictures that are highly consistent with the story and closely aligned with the intended narrative. To compare the results, the human utilizes the prompt from $M^{2}Chat$ to generate four pictures using the SDXL method. The blue blocks assess and contrast the images produced.}
  \label{fig:exp_story}
\end{figure*}

\subsection{Experiment Setup}
\label{sec:exp_setup}
\paragraph{Datasets.}
To minimize the domain gap between LLaMA-AdapterV2~\cite{gao2023llamaadapterv2} and SDXL~\cite{podell2023sdxl}, we aim to align them by using the shared dataset that they are familiar with. We tuned M$^{2}$Chat on the CC3M~\cite{sharma2018conceptual} and LAION-Aesthetics~\cite{schuhmann2022laion}, as shown in their technical report, the COCO-Caption dataset~\cite{lin2015microsoft} is also known for been used in training the VQA ability of VLM, which contains abundant objects descriptions. In terms of LAION-Aesthetics~\cite{schuhmann2022laion}, as a subset of LAION-5B~\cite{schuhmann2022laion}, shows good ability in improving the generalization quality. Then, we evaluate our data on the following datasets:
\begin{itemize}
    \item MS-COCO-Validation~\cite{lin2014microsoft} dataset is a carefully annotated subset of the MS-COCO dataset comprising diverse images. The dataset is generally used to assess the performance of computer vision models in tasks like object detection and segmentation.
    
    \item CC3M~\cite{sharma2018conceptual} (Conceptual Captions 3 Million) dataset is a vast collection of web-sourced images paired with natural language captions. It is aimed at advancing machine learning in image understanding and caption generation.
    
    \item MMDialog~\cite{feng2022mmdialog} dataset is a multimodal collection containing annotated dialogues with paired textual conversation and visual information. MMDialog is designed to facilitate research in multimodal dialogue systems.
\end{itemize}

\paragraph{Evaluation metrics.}
Our methodology undergoes evaluation through text-image generation metrics encompassing textual and visual dimensions. For visual assessment, we utilize CLIP-based metrics(\textbf{CLIP}) and Frechet Inception Distance (\textbf{FID}) to measure text-image congruence and image fidelity. Text analysis employs \textbf{BLEU}~\cite{10.3115/1073083.1073135} and \textbf{ROUGE}~\cite{lin-2004-rouge} . Specifically, we use \textbf{BLEU-1} and \textbf{BLEU-2}, which are variations of the BLEU. BLEU-1 assesses the accuracy of single-word correspondences in machine translation against reference texts, whereas BLEU-2 expands the evaluation to bigrams, enhancing the assessment of phrase translation quality. Moreover, as multimodal dialogue metrics require modality-specific precision, text-image relation, and contextual harmony, we introduce \textbf{InterRel} to overcome the traditional metrics' limitations by using CLIP to quantify conversational cross-modality alignment according to the MM-Relevance\cite{feng2022mmdialog}. For each interaction, let $\tilde{R}_T$ and $\tilde{R}_V$ be the generated text and image, and $R_T$ and $R_V$ are the ground truths. InterRel uses their CLIP embeddings, $\tilde{E}_M$ and $E_M$ for $M \in \{T,V\}$, to quantify relevance:
\begin{equation}
\begin{aligned}
\label{eq:m^{3}e}
\mathrm{InterRel}(R,\tilde{R}) = \sum_{i=0}(e_{i}^{M_1} \cdot \tilde{e}_{i}^{M_2} + e_{i}^{M_1} \cdot \tilde{e}_{i+1}^{M_2})
\end{aligned}
\end{equation}
where $e_{i}^{M_1} \in E_{M_1}$, $\tilde{e}_{i}^{M_2}, \tilde{e}_{i+1}^{M_2} \in \tilde{E}_{M_2}$,  $M_1 \neq M_2$ and $M_1, M_2  \in \{T,V\}$.

\begin{table*}[t]
\centering
\caption{Evaluation results based on FID and CLIP on CC3M and MS-COCO 2014 Validation set.}
\label{tab:clip}
\footnotesize
\setlength{\tabcolsep}{8mm}{
	\resizebox{1.99\columnwidth}{!}{
\begin{tabular}{lcccc}
\toprule
\multirow{2}{*}{Models} & \multicolumn{2}{c}{MS-COCO 2014} & \multicolumn{2}{c}{CC3M} \\
\cmidrule(r){2-3} \cmidrule(r){4-5}
\multirow{2}{*}{} & LLM Size & CLIP $\uparrow$ & FID $\downarrow$  & CLIP $\uparrow$\\ 
\midrule \midrule
SD 1.5                            & - & 30.62 & 30.62 & 23.48 \\
SDXL~\cite{podell2023sdxl}        & - &  {31.17} &  {24.26} &  {29.91} \\
\midrule
Emu ~\cite{sun2023generative} & 13B       & 28.6    & - & -\\
Emu2-Gen ~\cite{Emu2}  & 33B       & 29.7    & - & -\\
NeXT-GPT ~\cite{wu2023nextgpt}  & 7B       & 29.31    & - & -\\
MiniGPT5~\cite{zheng2023minigpt5} & 7B & -     & 31.47 & 22.00\\
\midrule
\textbf{M$^2$Chat} & 7B & 28.46 & 28.71 & 21.87\\ 
\textbf{M$^2$Chat (M$^3$FT)} & 7B &  {29.87} &  {26.15} &  {23.51}\\ 
\bottomrule
\end{tabular}
}}
\end{table*}
\begin{figure}[t]
\includegraphics[width=0.95\linewidth]{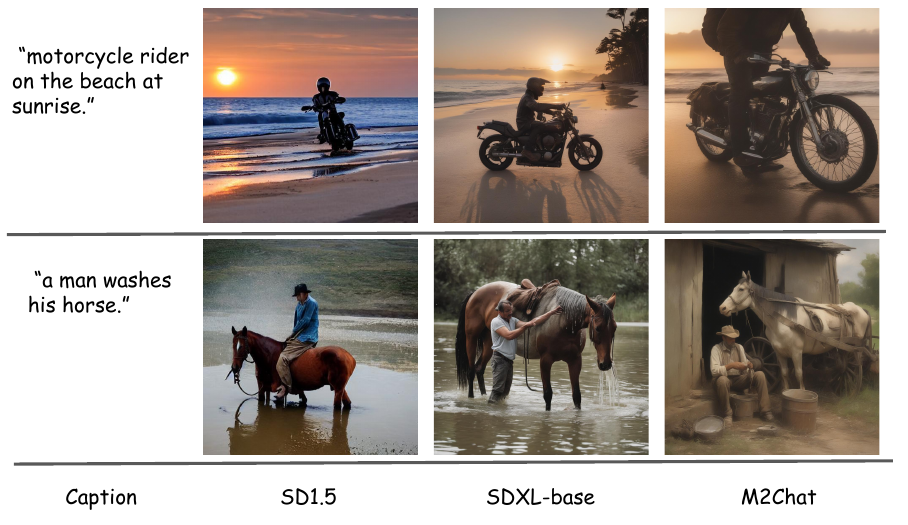}
\caption{Generation performance comparison of one-stage M$^{2}$Chat with SD1.5 and SDXL in Text-to-Image generation task. }
\label{fig:t2i}
\end{figure}

\paragraph{Baselines.}
To thoroughly assess our multimodal generation capabilities, we performed comparative analyses against multiple state-of-the-art baseline models targeting different perspectives: Stable Diffusion 1.5 and its larger counterpart, SDXL~\cite{podell2023sdxl}, are cutting-edge AI models generating detailed and varied images from text, with SDXL delivering superior fidelity; Emu~\cite{sun2023generative} introduces quality-tuning to enhance pre-trained models for generating compelling visuals without compromising concept versatility, whereas its successor, Emu2~\cite{Emu2}, advances training on extensive multimodal sequences under a unified autoregressive framework. SEED-LLaMA~\cite{ge2023seed2} enhances LLMs with an image tokenizer for synchronous text-image generation; NExT-GPT~\cite{wu2023nextgpt} integrates an LLM with multimodal adaptors and diverse diffusion decoders, endowing NExT-GPT with the capacity to process and output multimodal data, while DreamLLM~\cite{dong2023dreamllm} and MiniGPT5~\cite{zheng2023minigpt5} have been introduced in Sec.~\ref{sec:intro}.

\paragraph{Implementaotin Details.}
Our model was trained end-to-end on eight H800 GPUs. As illustrated in Fig.~\ref{fig:framework}, we focused on training the M$^{3}$Adapter exclusively. The VLM backbone, LLaMA-AdapterV2 7B, was paired with CLIP(ViT-L/14)\cite{radford2021clip} for visual encoding. The M$^{3}$Adapter's parameters occupy 299Mb, with an inference memory of 28Gb. 
During the First Stage in M$^3$FT for Alignment, we initialized a learning rate of 1e$^{-4}$, a batch size of 8, and conducted over 4 epochs, the training required approximately 80 GPU hours in total. We trained on a subset of CC3M\cite{sharma2018conceptual} with around 1.5 million image-text pairs. 

\begin{table*}[t]
\centering
\caption{Evaluation results of InterRel on MMDialog Validation set. }
\label{tab:mmdialog}
\footnotesize
\setlength{\tabcolsep}{4mm}{
	\resizebox{1.99\columnwidth}{!}{
\begin{tabular}{lccccc}
\toprule
Models & LLM & BLEU-1$\uparrow$ &BLEU-2$\uparrow$ & ROUGE-L$\uparrow$ & InterRel$\uparrow$ \\ 
\midrule \midrule
VLM+SD finetune & Vicuna 7B & 4.21 & 4.18 & 6.78 & 20.05 \\
\midrule
M$^2$Chat & LLaMA 7B & 6.02 & 5.88 & 10.14 & 24.68\\
M$^2$Chat(M$^3FT$) & LLaMA 7B & 6.98 & 6.44 & 11.40 & 25.57\\
\bottomrule
\end{tabular}
}
}
\end{table*}

\begin{figure*}[t]
  \centering
    \includegraphics[width=0.95\linewidth]{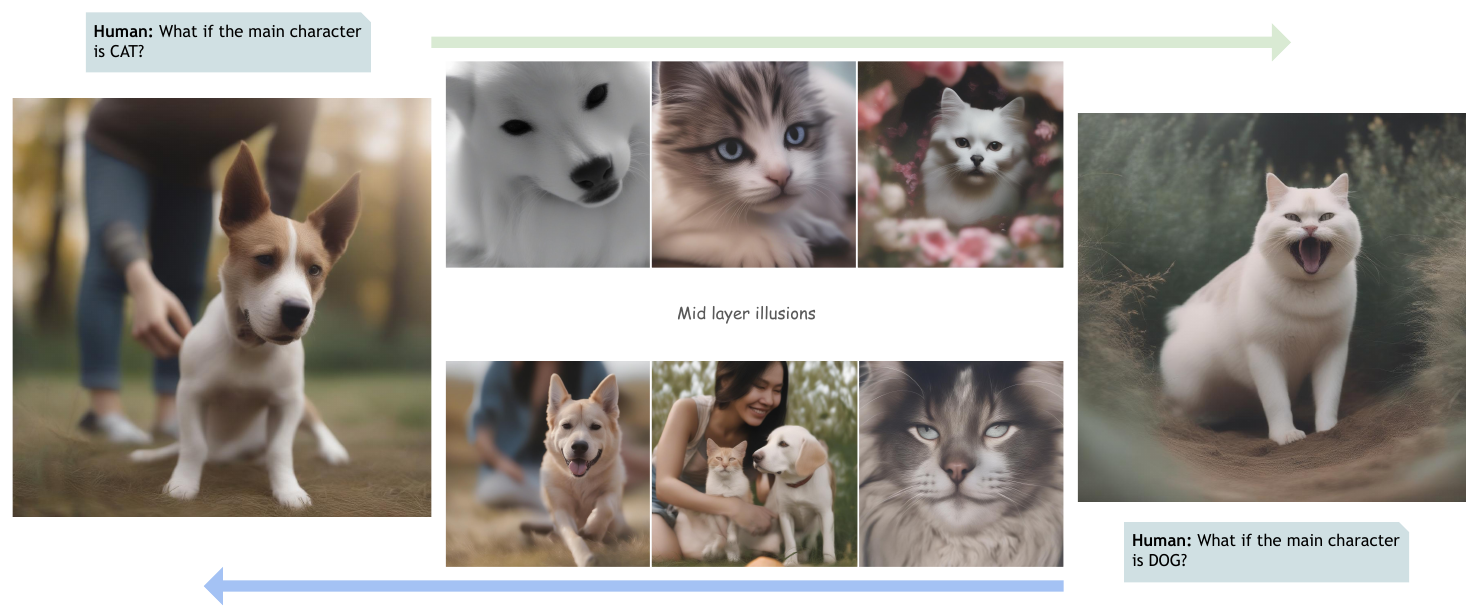}
  \caption{Visualization of the transformation of the hidden features while doing the instruction editing task. Giving the Dog picture and human instruction, the hidden features of VLM gradually transform its representation from dog to cat. The opposite instruction, which turns the cat into a dog, also shows a similar transformation step.}
  \label{fig:hidden_t2i}
\end{figure*}

During the second stage in M$^3$FT for Alignment, we initialized a learning rate of 1e$^{-5}$, a batch size of 1, and conducted over 20 epochs, the training required approximately 30 GPU hours in total. We train all the adapters by a mixture dataset, with 4k image-text instruction paired data extracted from CC3M, and 7k MMdialog conversation pairs from the training set of MMDialog\cite{feng2022mmdialog}. The learning rate is initialized at 1e-4, and decays 10 times each five epochs.
\subsection{Quantitative Results}

In our evaluation, we conducted a performance comparison of our model, $M^{2}Chat$, on the MS-COCO 2014 and CC3M validation datasets, as outlined in Table \ref{tab:clip}. Our results demonstrate the competitive performance of $M^{2}Chat$ compared to other generative models.
\paragraph{MS-COCO dataset} Our model achieves a state-of-the-art (SOTA) score of 29.87, surpassing other multimodality generation models by a margin of 0.56. The score also notably outperforms NExT-GPT \cite{wu2023nextgpt}, and slightly surpasses the large-scale pre-trained model Emu2 \cite{Emu2}.
\paragraph{CC3M validation set} We compared our results with MiniGPT5 \cite{zheng2023minigpt5}, which has a similar-sized LLM to M$^2$Chat. Our model demonstrates superior performance, achieving a 2.56 improvement in the FID score and a 1.51 improvement in the CLIP score.
\paragraph{MMDialog} We conducted a comparison of our results with a baseline model, VLM+SD finetune, which served as a reference point.Both our model, M$^2$Chat, and the baseline model followed the same pretraining and finetuning settings. The results demonstrate the effectiveness of our alignment method, as evidenced by the significant improvements in various evaluation metrics. For instance, our model achieved a remarkable 5.52 increase in the InterRel score. Additionally, there are notable improvements of 2.77 in BLEU-1, 2.16 in BLEU-2, and 4.62 in ROUGE-L scores.  It is important to note that the base multimodality model used in this baseline, LlaMA-AdapterV2, is not specifically fine-tuned for chat applications, which leads to relatively lower language scores.



\subsection{Qualitative Comparisons}
\label{sec:qc}
\paragraph{Image Generation Quality}
As shown in Fig.~\ref{fig:t2i}, our pipeline generates high-resolution images in different contents. It is demonstrated that our efficient alignment methods adapt the prompts well, as described in the quantitative results. Our $M^{2}Chat$ without M$^3$FT is compared with SDXL-base and SD1.5 for a fair comparison. Here, the generalization resolution is 1024$\times$1024. In conclusion, M$^2$Chat is able to fit the prompt better than SD1.5. We provide more generation results in Appendix A.

\paragraph{Storytelling}
We show the storytelling ability of M$^2$Chat on Fig.~\ref{fig:exp_story}. While asking the M$^2$Chat to tell us a story, it generates a story composed of text together with four pictures that follow the storyline. In comparison, we made a set of pictures that were artificially produced: fix the random seed of the SDXL, use the prompts generated by $M^{2}Chat$, and feed it to the SDXL. Our method shows high consistency of the text-images among multi-turn conversations. $M^{2}Chat$ performs better in showing the progress of the story, especially in the last two pictures. It shows the progress from "defend himself " to "the dragon died" since the SDXL is limited by the prompt size. We provide more comprehensive generation results in Appendix C.

\begin{figure}[t]
  \centering
    \includegraphics[width=0.95\linewidth]{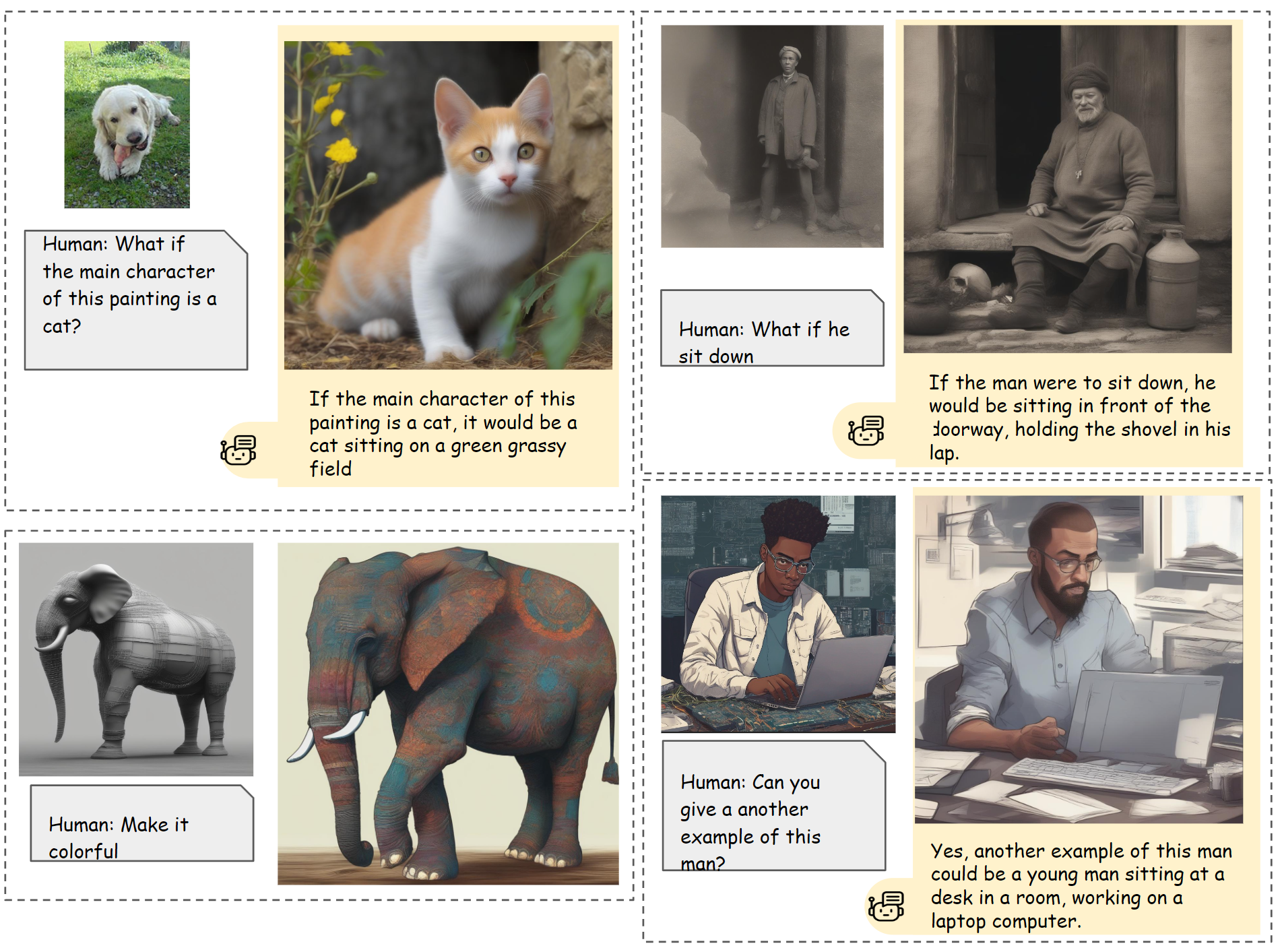}
  \caption{Examples of interleaved zero-shot image editing. M$^2$Chat consistently demonstrates excellent representation consistency while adhering to the editing instructions.}
  \label{fig:exp_edit}
\end{figure}

\paragraph{Interleaved editing}
Interleaved zero-shot image editing refers to the process of modifying images based on textual instructions without the need for paired image-text data during training. The goal is to achieve consistent and accurate image editing results by leveraging the learned representations from a pre-trained model. M$^2$Chat consistently demonstrates excellent representation consistency while faithfully adhering to the editing instructions. As shown in the Fig.~\ref{fig:exp_edit}, M$^2$Chat can edit the pose, replace the character, give a similar picture, etc.

\begin{table*}[t]
\centering
\caption{Comparison of parameter size and training cost with other multimodality generation models}
\label{tab:parameter}
\footnotesize
\resizebox{1.99\columnwidth}{!}{
\begin{tabular}{lcccccc}
\toprule
Models & LLM & Extra parameter & Data scale & Task & Wall-clock time\\ 
\midrule \midrule
Emu2~\cite{Emu2} & LLaMA 33B & 4B & 100M & TI $\rightarrow$ TI & - \\
CAFE~\cite{zhou2023customization} & LLaMA 70B & 4B & - & TI $\rightarrow$ TI & 20000$\times$A100 Hrs \\
SEED-LLaMA-8B~\cite{ge2023seed2} & Vicuna 7B & 1B & - & TI $\rightarrow$ TI & 9000 A100(40G) Hrs \\
SEED-LLaMA-14B~\cite{ge2023seed2} & LLaMA 13B & 1B & - & TI $\rightarrow$ TI & 14000 A100(40G) Hrs \\
DreamLLM~\cite{dong2023dreamllm} & Vicuna 7B & - & 32M & TI $\rightarrow$ TI & 2240 A100 Hrs \\
MiniGPT5~\cite{zheng2023minigpt5} & Vicuna 7B & - & 2.5M & TI $\rightarrow$ TI & - \\
\midrule
\textbf{M$^2$Chat} & LLaMA 7B & \textbf{299M} & \textbf{2M} & TI $\rightarrow$ TI & \textbf{100 A100 Hrs}\\ 
\bottomrule
\end{tabular}
}
\end{table*}

\paragraph{Multi-level feature visualization}
As previously mentioned in Sec.~\ref{method}, we employed multi-level feature fusion in our approach. Additionally, we visualized the hidden layer features of the LLM. In Fig.~\ref{fig:hidden_t2i}, we presented the process wherein M$^{2}$Chat effectively adheres to given instructions, resulting in the transformation of the dog depicted in the left image into a cat in the corresponding right image. The model takes human instruction and the picture as input, and outputs the image captions as well as the edited pictures. Furthermore, as shown in the Fig.~\ref{fig:teaser}, M$^{2}$Chat also supports multi-round editing. More results will be shown in Appendix B.


\subsection{Ablation Study}
\paragraph{Ablation of M$^{3}$FT}
In this paper, we claim that the low-level visual information and high-level semantic features have different effects on the final generalization. Hidden layers inside the VLM contain different levels of information and show a strong tendency to transition from the given state to the output state. To visualize the difference between layers, the Fig.~\ref{fig:hidden_t2i} shows the visualization of middle layers in the text-image and image-text tasks. 
Furthermore, as shown in the Tab.~\ref{tab:clip}, we compare the M$^2$Chat with the no M$^3$FT version. With the M$^3$FT, the CLIP score of MS-COCO improves by 1.39. On CC3M dataset, the M$^3$FT improves 2.56 in FID score, and 1.64 in the CLIP score. Both qualitative results and quantitative results illustrate the importance and efficiency of M$^3$FT.

\subsection{Efficiency Comparison}
Inspire by a serial of finetuning methods\cite{peft,zhang2024vecaf}, in Table \ref{tab:parameter}, we present a comparison of the training costs between our model and other multimodality and multitask generation models. The results clearly demonstrate that M$^2$Chat outperforms the other methods in terms of parameter efficiency and low training costs. For instance, Emu \cite{sun2023generative} and SEED \cite{ge2023seed1} focus on training large multimodality models without fully utilizing the potential of pretrained components, resulting in training costs exceeding 10,000 GPU hours. Similarly, DreamLLM \cite{dong2023dreamllm} incorporates learnable tokens to fine-tune the LLM for both understanding and generalization abilities, which incurs training costs exceeding 2,240 GPU hours.

In comparison, M$^2$Chat demonstrates a close data scale to MiniGPT5, around 2.5 million. Moreover, the additional parameters in M$^2$Chat are highly efficient when compared to the billion-level parameters found in other works.
\section{Conclusion}
In this paper, we introduce a novel multimodal interleaved text-image generation framework called \textit{$M^{2}Chat$}, which capables of generating text and images simultaneously. M$^{2}$Chat is built upon the VLM LLaMA-AdapterV2, with SDXL. We leverage a lightweight module M$^{3}$Adapter to achieve multimodal feature alignment. Moreover,  we further integrate the low-level feature with high-level features via a innovative gating strategy to balance the model creativity and coherence. Last but not least, we propose a two-stage M$^{3}$FT to further enhance semantic consistency. Extensive experiments demonstrate the superiority of M$^{2}$Chat across various multimodal interleaved tasks.

{
    \small
    \bibliographystyle{ieeenat_fullname}
    \bibliography{main}

\begin{thebibliography}{49}
\providecommand{\natexlab}[1]{#1}
\providecommand{\url}[1]{\texttt{#1}}
\expandafter\ifx\csname urlstyle\endcsname\relax
  \providecommand{\doi}[1]{doi: #1}\else
  \providecommand{\doi}{doi: \begingroup \urlstyle{rm}\Url}\fi

\bibitem[Cao et~al.(2023)Cao, Wang, Qi, Shan, Qie, and Zheng]{cao2023masactrl}
Mingdeng Cao, Xintao Wang, Zhongang Qi, Ying Shan, Xiaohu Qie, and Yinqiang Zheng.
\newblock Masactrl: Tuning-free mutual self-attention control for consistent image synthesis and editing.
\newblock \emph{arXiv preprint arXiv:2304.08465}, 2023.

\bibitem[Dong et~al.(2023)Dong, Han, Peng, Qi, Ge, Yang, Zhao, Sun, Zhou, Wei, et~al.]{dong2023dreamllm}
Runpei Dong, Chunrui Han, Yuang Peng, Zekun Qi, Zheng Ge, Jinrong Yang, Liang Zhao, Jianjian Sun, Hongyu Zhou, Haoran Wei, et~al.
\newblock Dreamllm: Synergistic multimodal comprehension and creation.
\newblock \emph{arXiv preprint arXiv:2309.11499}, 2023.

\bibitem[Feng et~al.(2022)Feng, Sun, Xu, Zhao, Yang, Tao, Zhao, and Lin]{feng2022mmdialog}
Jiazhan Feng, Qingfeng Sun, Can Xu, Pu Zhao, Yaming Yang, Chongyang Tao, Dongyan Zhao, and Qingwei Lin.
\newblock Mmdialog: A large-scale multi-turn dialogue dataset towards multi-modal open-domain conversation.
\newblock \emph{arXiv preprint arXiv:2211.05719}, 2022.

\bibitem[Gao et~al.(2023)Gao, Han, Zhang, Lin, Geng, Zhou, Zhang, Lu, He, Yue, Li, and Qiao]{gao2023llamaadapterv2}
Peng Gao, Jiaming Han, Renrui Zhang, Ziyi Lin, Shijie Geng, Aojun Zhou, Wei Zhang, Pan Lu, Conghui He, Xiangyu Yue, Hongsheng Li, and Yu Qiao.
\newblock Llama-adapter v2: Parameter-efficient visual instruction model.
\newblock \emph{arXiv preprint arXiv:2304.15010}, 2023.

\bibitem[Ge et~al.(2023{\natexlab{a}})Ge, Ge, Zeng, Wang, and Shan]{ge2023seed2}
Yuying Ge, Yixiao Ge, Ziyun Zeng, Xintao Wang, and Ying Shan.
\newblock Planting a seed of vision in large language model.
\newblock \emph{arXiv preprint arXiv:2307.08041}, 2023{\natexlab{a}}.

\bibitem[Ge et~al.(2023{\natexlab{b}})Ge, Zhao, Zeng, Ge, Li, Wang, and Shan]{ge2023seed1}
Yuying Ge, Sijie Zhao, Ziyun Zeng, Yixiao Ge, Chen Li, Xintao Wang, and Ying Shan.
\newblock Making llama see and draw with seed tokenizer.
\newblock \emph{arXiv preprint arXiv:2310.01218}, 2023{\natexlab{b}}.

\bibitem[Hertz et~al.(2022)Hertz, Mokady, Tenenbaum, Aberman, Pritch, and Cohen-Or]{hertz2022prompt}
Amir Hertz, Ron Mokady, Jay Tenenbaum, Kfir Aberman, Yael Pritch, and Daniel Cohen-Or.
\newblock Prompt-to-prompt image editing with cross attention control.
\newblock \emph{arXiv preprint arXiv:2208.01626}, 2022.

\bibitem[Huang et~al.(2023)Huang, Dong, Wang, Hao, Singhal, Ma, Lv, Cui, Mohammed, Liu, et~al.]{huang2023language}
Shaohan Huang, Li Dong, Wenhui Wang, Yaru Hao, Saksham Singhal, Shuming Ma, Tengchao Lv, Lei Cui, Owais~Khan Mohammed, Qiang Liu, et~al.
\newblock Language is not all you need: Aligning perception with language models.
\newblock \emph{arXiv preprint arXiv:2302.14045}, 2023.

\bibitem[Koh et~al.(2023{\natexlab{a}})Koh, Fried, and Salakhutdinov]{koh2023generating}
Jing~Yu Koh, Daniel Fried, and Ruslan Salakhutdinov.
\newblock Generating images with multimodal language models.
\newblock \emph{arXiv preprint arXiv:2305.17216}, 2023{\natexlab{a}}.

\bibitem[Koh et~al.(2023{\natexlab{b}})Koh, Salakhutdinov, and Fried]{koh2023grounding}
Jing~Yu Koh, Ruslan Salakhutdinov, and Daniel Fried.
\newblock Grounding language models to images for multimodal generation.
\newblock \emph{arXiv preprint arXiv:2301.13823}, 2023{\natexlab{b}}.

\bibitem[Lai et~al.(2023)Lai, Zhu, Dai, Qiao, and Wang]{minidalle3}
Zeqiang Lai, Xizhou Zhu, Jifeng Dai, Yu Qiao, and Wenhai Wang.
\newblock Mini-dalle3: Interactive text to image by prompting large language models, 2023.

\bibitem[Li et~al.(2023)Li, Li, Savarese, and Hoi]{li2023blip}
Junnan Li, Dongxu Li, Silvio Savarese, and Steven Hoi.
\newblock Blip-2: Bootstrapping language-image pre-training with frozen image encoders and large language models.
\newblock \emph{arXiv preprint arXiv:2301.12597}, 2023.

\bibitem[Lian et~al.(2023)Lian, Li, Yala, and Darrell]{lian2023llm}
Long Lian, Boyi Li, Adam Yala, and Trevor Darrell.
\newblock Llm-grounded diffusion: Enhancing prompt understanding of text-to-image diffusion models with large language models.
\newblock \emph{arXiv preprint arXiv:2305.13655}, 2023.

\bibitem[Lin(2004)]{lin-2004-rouge}
Chin-Yew Lin.
\newblock {ROUGE}: A package for automatic evaluation of summaries.
\newblock In \emph{Text Summarization Branches Out}, pages 74--81, Barcelona, Spain, 2004. Association for Computational Linguistics.

\bibitem[Lin et~al.(2014)Lin, Maire, Belongie, Hays, Perona, Ramanan, Doll{\'a}r, and Zitnick]{lin2014microsoft}
Tsung-Yi Lin, Michael Maire, Serge Belongie, James Hays, Pietro Perona, Deva Ramanan, Piotr Doll{\'a}r, and C~Lawrence Zitnick.
\newblock Microsoft coco: Common objects in context.
\newblock In \emph{Computer Vision--ECCV 2014: 13th European Conference, Zurich, Switzerland, September 6-12, 2014, Proceedings, Part V 13}, pages 740--755. Springer, 2014.

\bibitem[Lin et~al.(2015)Lin, Maire, Belongie, Bourdev, Girshick, Hays, Perona, Ramanan, Zitnick, and Dollár]{lin2015microsoft}
Tsung-Yi Lin, Michael Maire, Serge Belongie, Lubomir Bourdev, Ross Girshick, James Hays, Pietro Perona, Deva Ramanan, C.~Lawrence Zitnick, and Piotr Dollár.
\newblock Microsoft coco: Common objects in context, 2015.

\bibitem[Lin et~al.(2023)Lin, Liu, Zhang, Gao, Qiu, Xiao, Qiu, Lin, Shao, Chen, et~al.]{lin2023sphinx}
Ziyi Lin, Chris Liu, Renrui Zhang, Peng Gao, Longtian Qiu, Han Xiao, Han Qiu, Chen Lin, Wenqi Shao, Keqin Chen, et~al.
\newblock Sphinx: The joint mixing of weights, tasks, and visual embeddings for multi-modal large language models.
\newblock \emph{arXiv preprint arXiv:2311.07575}, 2023.

\bibitem[Liu et~al.(2023{\natexlab{a}})Liu, Li, Li, and Lee]{liu2023improvedllava}
Haotian Liu, Chunyuan Li, Yuheng Li, and Yong~Jae Lee.
\newblock Improved baselines with visual instruction tuning, 2023{\natexlab{a}}.

\bibitem[Liu et~al.(2023{\natexlab{b}})Liu, Li, Wu, and Lee]{liu2023llava}
Haotian Liu, Chunyuan Li, Qingyang Wu, and Yong~Jae Lee.
\newblock Visual instruction tuning, 2023{\natexlab{b}}.

\bibitem[Liu et~al.(2024{\natexlab{a}})Liu, Li, Li, Li, Zhang, Shen, and Lee]{liu2024llava}
Haotian Liu, Chunyuan Li, Yuheng Li, Bo Li, Yuanhan Zhang, Sheng Shen, and Yong~Jae Lee.
\newblock Llava-next: Improved reasoning, ocr, and world knowledge, january 2024a.
\newblock \emph{URL https://llava-vl. github. io/blog/2024-01-30-llava-next}, 2024{\natexlab{a}}.

\bibitem[Liu et~al.(2024{\natexlab{b}})Liu, Li, Wu, and Lee]{liu2024visual}
Haotian Liu, Chunyuan Li, Qingyang Wu, and Yong~Jae Lee.
\newblock Visual instruction tuning.
\newblock \emph{Advances in neural information processing systems}, 36, 2024{\natexlab{b}}.

\bibitem[Liu et~al.(2023{\natexlab{c}})Liu, Li, Li, Yu, Huang, Peng, Liu, Chen, Li, Jin, et~al.]{liu2023hidden}
Yuliang Liu, Zhang Li, Hongliang Li, Wenwen Yu, Mingxin Huang, Dezhi Peng, Mingyu Liu, Mingrui Chen, Chunyuan Li, Lianwen Jin, et~al.
\newblock On the hidden mystery of ocr in large multimodal models.
\newblock \emph{arXiv preprint arXiv:2305.07895}, 2023{\natexlab{c}}.

\bibitem[Mangrulkar et~al.(2022)Mangrulkar, Gugger, Debut, Belkada, Paul, and Bossan]{peft}
Sourab Mangrulkar, Sylvain Gugger, Lysandre Debut, Younes Belkada, Sayak Paul, and Benjamin Bossan.
\newblock Peft: State-of-the-art parameter-efficient fine-tuning methods.
\newblock \url{https://github.com/huggingface/peft}, 2022.

\bibitem[Nichol and Dhariwal(2021)]{nichol2021improved}
Alexander~Quinn Nichol and Prafulla Dhariwal.
\newblock Improved denoising diffusion probabilistic models.
\newblock In \emph{International Conference on Machine Learning}, pages 8162--8171. PMLR, 2021.

\bibitem[OpenAI(2023)]{openai2023dalle3}
OpenAI.
\newblock Improving image generation with better captions, 2023.

\bibitem[Papineni et~al.(2002)Papineni, Roukos, Ward, and Zhu]{10.3115/1073083.1073135}
Kishore Papineni, Salim Roukos, Todd Ward, and Wei-Jing Zhu.
\newblock Bleu: a method for automatic evaluation of machine translation.
\newblock In \emph{Proceedings of the 40th Annual Meeting on Association for Computational Linguistics}, page 311–318, USA, 2002. Association for Computational Linguistics.

\bibitem[Podell et~al.(2023)Podell, English, Lacey, Blattmann, Dockhorn, M{\"u}ller, Penna, and Rombach]{podell2023sdxl}
Dustin Podell, Zion English, Kyle Lacey, Andreas Blattmann, Tim Dockhorn, Jonas M{\"u}ller, Joe Penna, and Robin Rombach.
\newblock Sdxl: Improving latent diffusion models for high-resolution image synthesis.
\newblock \emph{arXiv preprint arXiv:2307.01952}, 2023.

\bibitem[Qi et~al.(2023)Qi, Pan, Li, Yuan, Chi, Li, Luo, Xue, Zhang, Liu, et~al.]{qi2023weakly}
Xingqun Qi, Jiahao Pan, Peng Li, Ruibin Yuan, Xiaowei Chi, Mengfei Li, Wenhan Luo, Wei Xue, Shanghang Zhang, Qifeng Liu, et~al.
\newblock Weakly-supervised emotion transition learning for diverse 3d co-speech gesture generation.
\newblock \emph{arXiv preprint arXiv:2311.17532}, 2023.

\bibitem[Qu et~al.(2023)Qu, Wu, Fei, Nie, and Chua]{qu2023layoutllm}
Leigang Qu, Shengqiong Wu, Hao Fei, Liqiang Nie, and Tat-Seng Chua.
\newblock Layoutllm-t2i: Eliciting layout guidance from llm for text-to-image generation.
\newblock In \emph{Proceedings of the 31st ACM International Conference on Multimedia}, pages 643--654, 2023.

\bibitem[Radford et~al.(2021)Radford, Kim, Hallacy, Ramesh, Goh, Agarwal, Sastry, Askell, Mishkin, Clark, et~al.]{radford2021clip}
Alec Radford, Jong~Wook Kim, Chris Hallacy, Aditya Ramesh, Gabriel Goh, Sandhini Agarwal, Girish Sastry, Amanda Askell, Pamela Mishkin, Jack Clark, et~al.
\newblock Learning transferable visual models from natural language supervision.
\newblock In \emph{International conference on machine learning}, pages 8748--8763. PMLR, 2021.

\bibitem[Schuhmann et~al.(2022)Schuhmann, Beaumont, Vencu, Gordon, Wightman, Cherti, Coombes, Katta, Mullis, Wortsman, et~al.]{schuhmann2022laion}
Christoph Schuhmann, Romain Beaumont, Richard Vencu, Cade Gordon, Ross Wightman, Mehdi Cherti, Theo Coombes, Aarush Katta, Clayton Mullis, Mitchell Wortsman, et~al.
\newblock Laion-5b: An open large-scale dataset for training next generation image-text models.
\newblock \emph{Advances in Neural Information Processing Systems}, 35:\penalty0 25278--25294, 2022.

\bibitem[Sharma et~al.(2018)Sharma, Ding, Goodman, and Soricut]{sharma2018conceptual}
Piyush Sharma, Nan Ding, Sebastian Goodman, and Radu Soricut.
\newblock Conceptual captions: A cleaned, hypernymed, image alt-text dataset for automatic image captioning.
\newblock In \emph{Proceedings of the 56th Annual Meeting of the Association for Computational Linguistics (Volume 1: Long Papers)}, pages 2556--2565, 2018.

\bibitem[Song et~al.(2023)Song, Chai, Wang, Zhang, Zhou, Wu, Guo, Ye, Lu, Hwang, et~al.]{song2023moviechat}
Enxin Song, Wenhao Chai, Guanhong Wang, Yucheng Zhang, Haoyang Zhou, Feiyang Wu, Xun Guo, Tian Ye, Yan Lu, Jenq-Neng Hwang, et~al.
\newblock Moviechat: From dense token to sparse memory for long video understanding.
\newblock \emph{arXiv preprint arXiv:2307.16449}, 2023.

\bibitem[Song et~al.(2020{\natexlab{a}})Song, Sohl-Dickstein, Kingma, Kumar, Ermon, and Poole]{song2020score}
Yang Song, Jascha Sohl-Dickstein, Diederik~P Kingma, Abhishek Kumar, Stefano Ermon, and Ben Poole.
\newblock Score-based generative modeling through stochastic differential equations.
\newblock \emph{arXiv preprint arXiv:2011.13456}, 2020{\natexlab{a}}.

\bibitem[Song et~al.(2020{\natexlab{b}})Song, Sohl-Dickstein, Kingma, Kumar, Ermon, and Poole]{Song2020ScoreBasedGM}
Yang Song, Jascha~Narain Sohl-Dickstein, Diederik~P. Kingma, Abhishek Kumar, Stefano Ermon, and Ben Poole.
\newblock Score-based generative modeling through stochastic differential equations.
\newblock \emph{ArXiv}, abs/2011.13456, 2020{\natexlab{b}}.

\bibitem[Sun et~al.(2023{\natexlab{a}})Sun, Han, Deng, Wang, Qin, and Gould]{sun20233dgpt}
Chunyi Sun, Junlin Han, Weijian Deng, Xinlong Wang, Zishan Qin, and Stephen Gould.
\newblock 3d-gpt: Procedural 3d modeling with large language models, 2023{\natexlab{a}}.

\bibitem[Sun et~al.(2023{\natexlab{b}})Sun, Cui, Zhang, Zhang, Yu, Luo, Wang, Rao, Liu, Huang, and Wang]{Emu2}
Quan Sun, Yufeng Cui, Xiaosong Zhang, Fan Zhang, Qiying Yu, Zhengxiong Luo, Yueze Wang, Yongming Rao, Jingjing Liu, Tiejun Huang, and Xinlong Wang.
\newblock Generative multimodal models are in-context learners.
\newblock \emph{arXiv preprint arXiv:2312.13286}, 2023{\natexlab{b}}.

\bibitem[Sun et~al.(2023{\natexlab{c}})Sun, Cui, Zhang, Zhang, Yu, Luo, Wang, Rao, Liu, Huang, et~al.]{sun2023generative}
Quan Sun, Yufeng Cui, Xiaosong Zhang, Fan Zhang, Qiying Yu, Zhengxiong Luo, Yueze Wang, Yongming Rao, Jingjing Liu, Tiejun Huang, et~al.
\newblock Generative multimodal models are in-context learners.
\newblock \emph{arXiv preprint arXiv:2312.13286}, 2023{\natexlab{c}}.

\bibitem[Tang et~al.(2023)Tang, Yang, Zhu, Zeng, and Bansal]{tang2023codi}
Zineng Tang, Ziyi Yang, Chenguang Zhu, Michael Zeng, and Mohit Bansal.
\newblock Any-to-any generation via composable diffusion.
\newblock \emph{arXiv preprint arXiv:2305.11846}, 2023.

\bibitem[Wang et~al.(2023)Wang, Lv, Yu, Hong, Qi, Wang, Ji, Yang, Zhao, Song, et~al.]{wang2023cogvlm}
Weihan Wang, Qingsong Lv, Wenmeng Yu, Wenyi Hong, Ji Qi, Yan Wang, Junhui Ji, Zhuoyi Yang, Lei Zhao, Xixuan Song, et~al.
\newblock Cogvlm: Visual expert for pretrained language models.
\newblock \emph{arXiv preprint arXiv:2311.03079}, 2023.

\bibitem[Wu et~al.(2023)Wu, Fei, Qu, Ji, and Chua]{wu2023nextgpt}
Shengqiong Wu, Hao Fei, Leigang Qu, Wei Ji, and Tat-Seng Chua.
\newblock Next-gpt: Any-to-any multimodal llm.
\newblock \emph{arXiv preprint arXiv:2309.05519}, 2023.

\bibitem[Zhang et~al.(2023{\natexlab{a}})Zhang, Li, and Bing]{damonlpsg2023videoLLaMA}
Hang Zhang, Xin Li, and Lidong Bing.
\newblock Video-llama: An instruction-tuned audio-visual language model for video understanding.
\newblock \emph{arXiv preprint arXiv:2306.02858}, 2023{\natexlab{a}}.

\bibitem[Zhang et~al.(2023{\natexlab{b}})Zhang, Rao, and Agrawala]{zhang2023adding}
Lvmin Zhang, Anyi Rao, and Maneesh Agrawala.
\newblock Adding conditional control to text-to-image diffusion models, 2023{\natexlab{b}}.

\bibitem[Zhang et~al.(2023{\natexlab{c}})Zhang, Chi, Liu, Zhang, Du, and Wang]{zhang2023unimodal}
Rongyu Zhang, Xiaowei Chi, Guiliang Liu, Wenyi Zhang, Yuan Du, and Fangxin Wang.
\newblock Unimodal training-multimodal prediction: Cross-modal federated learning with hierarchical aggregation.
\newblock \emph{arXiv preprint arXiv:2303.15486}, 2023{\natexlab{c}}.

\bibitem[Zhang et~al.(2023{\natexlab{d}})Zhang, Han, Liu, Gao, Zhou, Hu, Yan, Lu, Li, and Qiao]{zhang2023LLaMAadapter}
Renrui Zhang, Jiaming Han, Chris Liu, Peng Gao, Aojun Zhou, Xiangfei Hu, Shilin Yan, Pan Lu, Hongsheng Li, and Yu Qiao.
\newblock Llama-adapter: Efficient fine-tuning of language models with zero-init attention.
\newblock \emph{arXiv preprint arXiv:2303.16199}, 2023{\natexlab{d}}.

\bibitem[Zhang et~al.(2024)Zhang, Cai, Yang, Liu, Gudovskiy, Okuno, Nakata, Keutzer, Chang, Du, et~al.]{zhang2024vecaf}
Rongyu Zhang, Zefan Cai, Huanrui Yang, Zidong Liu, Denis Gudovskiy, Tomoyuki Okuno, Yohei Nakata, Kurt Keutzer, Baobao Chang, Yuan Du, et~al.
\newblock Vecaf: Vlm-empowered collaborative active finetuning with training objective awareness.
\newblock \emph{arXiv preprint arXiv:2401.07853}, 2024.

\bibitem[Zheng et~al.(2023)Zheng, He, and Wang]{zheng2023minigpt5}
Kaizhi Zheng, Xuehai He, and Xin~Eric Wang.
\newblock Minigpt-5: Interleaved vision-and-language generation via generative vokens.
\newblock \emph{arXiv preprint arXiv:2310.02239}, 2023.

\bibitem[Zhou et~al.(2023)Zhou, Zhang, Gu, and Sun]{zhou2023customization}
Yufan Zhou, Ruiyi Zhang, Jiuxiang Gu, and Tong Sun.
\newblock Customization assistant for text-to-image generation.
\newblock \emph{arXiv preprint arXiv:2312.03045}, 2023.

\bibitem[Zhu et~al.(2023)Zhu, Chen, Shen, Li, and Elhoseiny]{zhu2023minigpt4}
Deyao Zhu, Jun Chen, Xiaoqian Shen, Xiang Li, and Mohamed Elhoseiny.
\newblock Minigpt-4: Enhancing vision-language understanding with advanced large language models.
\newblock \emph{arXiv preprint arXiv:2304.10592}, 2023.

\end{thebibliography}
}


\end{document}